\documentclass[letterpaper]{article} 
\usepackage{aaai2026}  
\usepackage{times}  
\usepackage{helvet}  
\usepackage{courier}  
\usepackage[hyphens]{url}  
\usepackage{graphicx} 
\usepackage{natbib}  
\usepackage{caption} 
\usepackage{amssymb}

\usepackage{algorithm}
\usepackage{algorithmic}

\usepackage{newfloat}
\usepackage{listings}
\DeclareCaptionStyle{ruled}{labelfont=normalfont,labelsep=colon,strut=off} 
\floatstyle{ruled}
\newfloat{listing}{tb}{lst}{}
\floatname{listing}{Listing}
\title{AMS-IO-Bench and AMS-IO-Agent: Benchmarking and Structured Reasoning for Analog and Mixed-Signal Integrated Circuit Input/Output Design}

\author{
	Zhishuai Zhang\equalcontrib\textsuperscript{\rm 1},
	Xintian Li\equalcontrib\textsuperscript{\rm 2},
	Shilong Liu\textsuperscript{\rm 3},
	Aodong Zhang\textsuperscript{\rm 1},
	Lu Jie\textsuperscript{\rm 2},
	Nan Sun\textsuperscript{\rm 1}
}
\affiliations{
	\textsuperscript{\rm 1}Department of Electrical Engineering, Tsinghua University\\
	\textsuperscript{\rm 2}School of Integrated Circuits, Tsinghua University\\
	\textsuperscript{\rm 3}Princeton AI Lab, Princeton University\\
	arcadia16zzs@gmail.com
}

\usepackage{bibentry}

\begin{document}

\maketitle

\begin{abstract}
In this paper, we propose \textit{AMS-IO-Agent}, a domain-specialized LLM-based agent for structure-aware input/output (I/O) subsystem generation in analog and mixed-signal (AMS) integrated circuits (ICs). 
The central contribution of this work is a framework that connects natural language design intent with industrial-level AMS IC design deliverables.
\textit{AMS-IO-Agent} integrates two key capabilities: 
(1) a structured domain knowledge base that captures reusable constraints and design conventions; 
(2) design intent structuring, which converts ambiguous user intent into verifiable logic steps using JSON and Python as intermediate formats.
We further introduce AMS-IO-Bench, a benchmark for wirebond-packaged AMS I/O ring automation. 
On this benchmark, AMS-IO-Agent achieves over 70\% DRC+LVS pass rate and reduces design turnaround time from hours to minutes, outperforming the baseline LLM. 
Furthermore, an agent-generated I/O ring was fabricated and validated in a 28 nm CMOS tape-out, demonstrating the practical effectiveness of the approach in real AMS IC design flows.
To our knowledge, this is the first reported human-agent collaborative AMS IC design in which an LLM-based agent completes a nontrivial subtask with outputs directly used in silicon.
\end{abstract}

\begin{links}
    \link{Code}{https://github.com/Arcadia-1/AMS-IO-Agent}
    \link{Datasets}{https://github.com/Arcadia-1/AMS-IO-Bench}
\end{links}

\section{Introduction}

Input/output (I/O) subsystems are a fundamental component of analog and mixed-signal (AMS) integrated circuits (ICs), providing signal interfacing, power delivery, and electrostatic discharge (ESD) protection.
While digital I/O cells can typically be placed using scripts and routed via standard digital flows, the implementation of AMS I/O remains largely manual due to intricate, project-specific requirements.
These include diverse signal types, multiple power domains, power integrity constraints, sensitive analog signal routing requirements, and layout restrictions imposed by fabrication and packaging rules.
Simple scripting approaches lack the reasoning capability to handle such complexity, leaving much of the design effort to human engineers.

\begin{figure}[t]
	\centering
	\includegraphics[width=1\columnwidth]{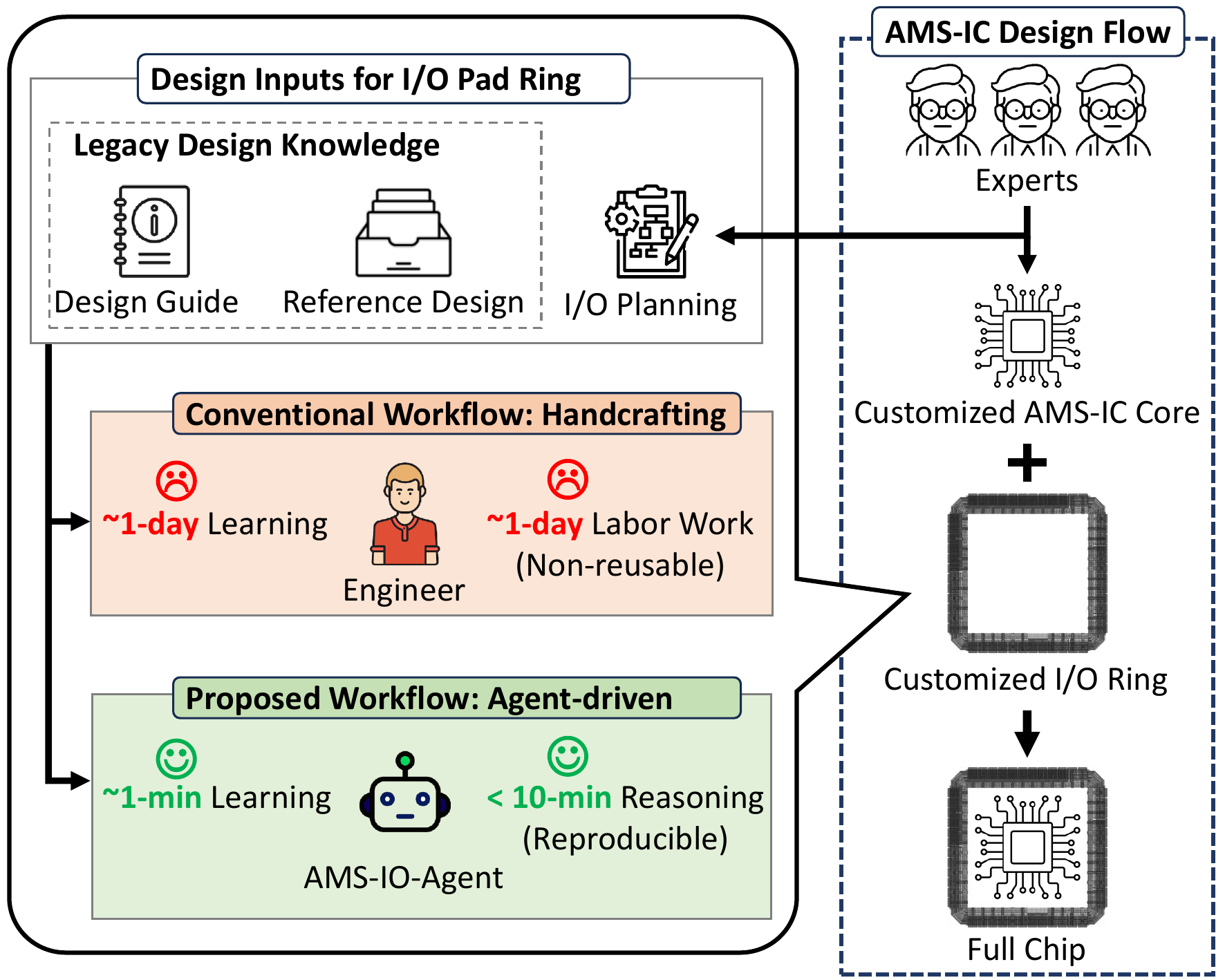} 
	\caption{Comparison of conventional and agent-driven I/O ring design in a real AMS IC design workflow.}
	\label{fig1}
\end{figure}

As a result, AMS I/O design is labor-intensive, with most effort rarely reusable. 
In wirebond-packaged chips (Fig.~\ref{fig1}), a novice engineer may spend one or two days on studying, manually assembling, and verifying I/O placement and connections.
Iterative pin changes often lead to disruptive rework risks, which intensify as tape-out approaches. 
These challenges highlight the need for intelligent automation, for which recent advances in large language models (LLMs) offer a promising foundation.

While LLMs have shown promising capabilities across several hardware design tasks \cite{Chen_2024, fang2025survey}, their application to AMS I/O design remains underexplored due to three key challenges: (1) the lack of accessible domain knowledge, which is typically confined to team-specific practices and scattered internal documents, (2) the absence of standardized task interfaces, as interaction still relies on GUI or domain-specific languages unfamiliar to pretrained LLMs, and (3) the unavailability of public benchmarks, which hinders systematic evaluation.

\begin{figure}[t]
	\centering
	\includegraphics[width=0.99\columnwidth]{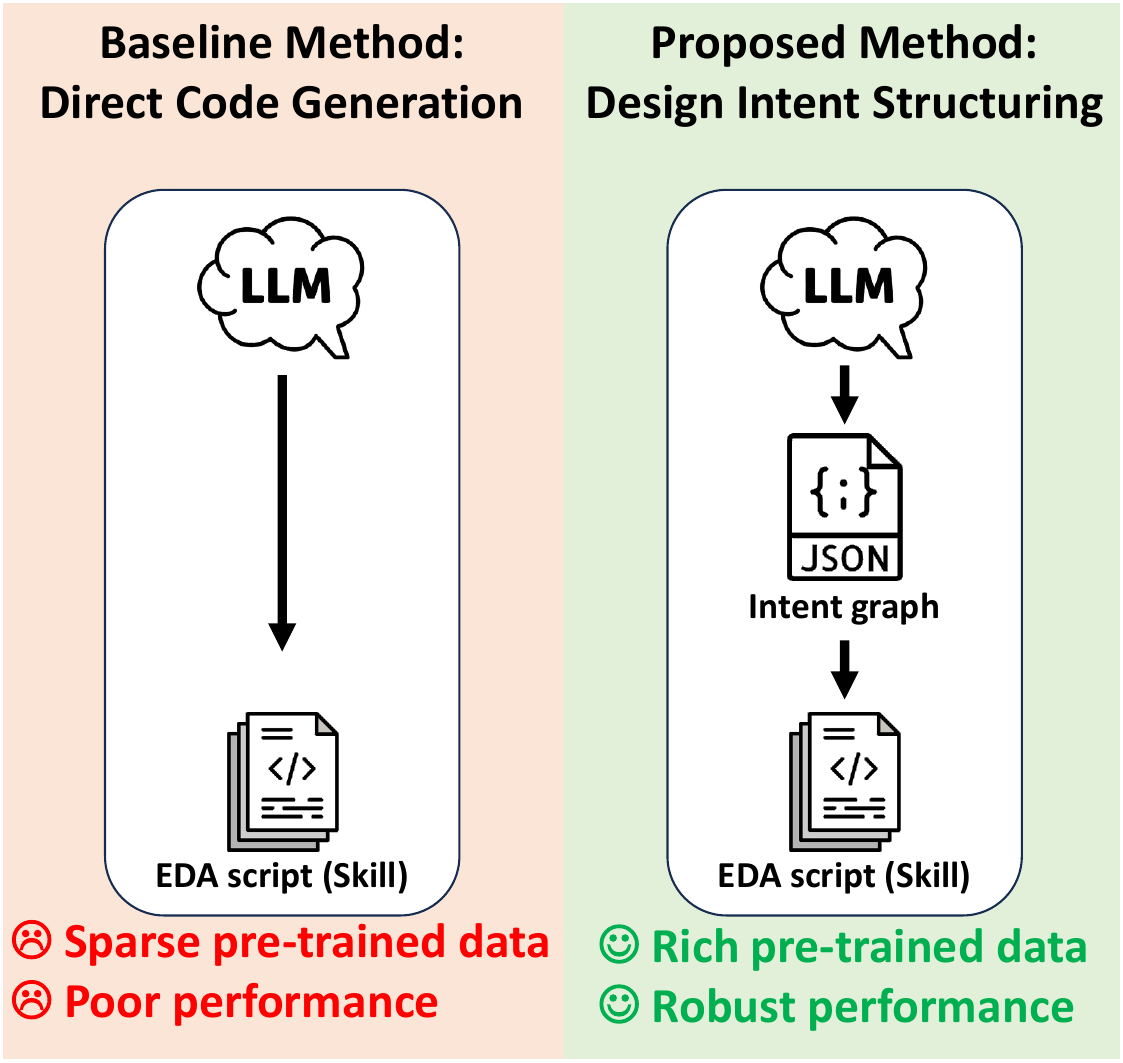} 
	\caption{Comparison of baseline direct code generation and the proposed structured design intent.}
	\label{fig:Comparison}
\end{figure}

To bridge this gap, we propose AMS-IO-Agent, a domain-specialized LLM-based agent for structure-aware AMS I/O generation.
It integrates two core capabilities:
(1) a domain knowledge base built from fragmented engineering practices, capturing reusable constraints and layout conventions, curated from real training materials developed by a professional AMS IC design team of more than 10 engineers and validated through over 50 successful tape-out cases;
(2) design intent structuring, which converts ambiguous design intent into verifiable logic steps as an intermediate format, as shown in Fig.~\ref{fig:Comparison}.
Together, these components allow the agent to adapt across projects by grounding generation in prior examples and domain constraints.

To support consistent evaluation, we also introduce AMS-IO-Bench, a benchmark for wirebond-packaged AMS I/O rings. 
It covers key tasks such as I/O ring assembly and validation, and is used to assess the correctness, adaptability, and efficiency of generated designs.
To the best of our knowledge, this is the first work to propose an LLM-based agent and comprehensive evaluation benchmark specifically targeting real-world AMS IC I/O ring design automation, and the first LLM-based agent for AMS IC design capable of direct integration into tape-out workflows as a key module collaborating with human engineers.

Experiments on AMS-IO-Bench show that AMS-IO-Agent achieves over 70\% DRC+LVS pass rates, surpassing the baseline LLM by a large margin and reducing design turnaround time from days to just minutes per case. Notably, we validated our system in real industrial tape-out projects: I/O rings generated by AMS-IO-Agent were seamlessly integrated into commercial AMS IC flows and successfully fabricated on silicon. This demonstrates not only the practicality and robustness of our approach, but also its readiness for deployment in real-world chip design pipelines.

In summary, our work makes the following contributions:
\begin{itemize}
	\item We propose an automatic pipeline for AMS IC I/O generation, formalizing the task into structured steps: intent interpretation, constraint resolution, and generation of EDA scripts, enabling practical automation.
	\item We introduce a novel agent architecture that combines domain-specific knowledge base and structured intent reasoning, supporting generalization across diverse AMS I/O design contexts.
	\item We develop a benchmark for wirebond-packaged AMS I/O ring automation, and show that the agent approach consistently delivers practical designs, reducing manual workload and design turnaround time.
	\item We further validate the agent in a real 28-nm CMOS tape-out, achieving the first reported demonstration of an LLM-based agent directly contributing to a  nontrivial AMS IC design task.
\end{itemize}

\section{Related Work}

\subsection{LLM-based Approaches for IC Design Automation}
LLMs have been applied to a variety of EDA tasks, from knowledge‑based question answering \cite{shi2024askedadesignassistantempowered, skelic2025circuitbenchmarkcircuitinterpretation} and register-transfer level (RTL) code generation \cite{chang2023chipgptfarnaturallanguage, Blocklove_2023, thakur2023verigenlargelanguagemodel, liu2023verilogevalevaluatinglargelanguage, xu2024llmaidedefficienthardwaredesign, 10691788, tsai2024rtlfixerautomaticallyfixingrtl, thakur2024autochipautomatinghdlgeneration, fu2025gpt4aigchipnextgenerationaiaccelerator} through netlist synthesis \cite{lai2024analogcoderanalogcircuitdesign}, circuit optimization \cite{Yin_2024, ghose2025orfsagenttoolusingagentschip}, to EDA script generation \cite{chen2025analogtesterlargelanguagemodelbased}.
To extend beyond one-shot code synthesis, recent studies adopt agent architectures that integrate LLMs into iterative design loops \cite{Wu_2024, ho2024largelanguagemodelllm, 10841395, liu2024chipnemodomainadaptedllmschip, ghose2025orfsagenttoolusingagentschip}. 
By invoking tools, observing results, and refining their outputs, these agents show promise for multi‑step workflows.

Nevertheless, these advances have not alleviated the repetitive, constraint-heavy routines in AMS IC design, as they lack systematic adaptation to real design flows and tape-out validation. As a result, constraint‑rich, project‑specific tasks remain largely manual, and the potential of LLMs to automate them is still untapped.

\begin{figure*}[t]
	\centering
	\includegraphics[width=0.9\textwidth]{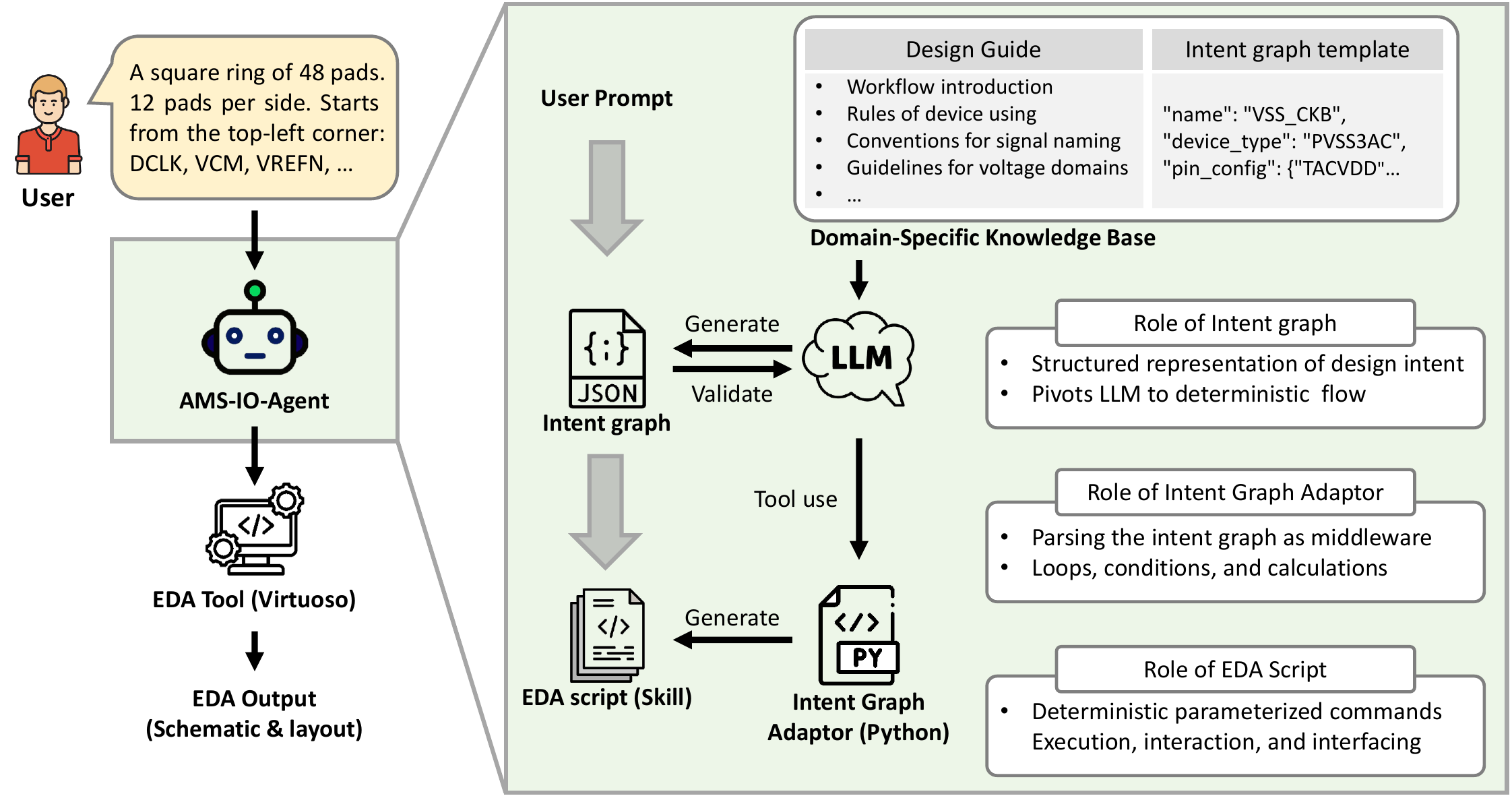} 
	\caption{Overview of the proposed AMS-IO-Agent architecture, illustrating the workflow and the component roles.}
	\label{fig:Overview}
\end{figure*}

\subsection{Benchmark for AMS IC agents}
The scarcity of domain-specific datasets and the inherent difficulty of precisely defining AMS IC design tasks make quantitative evaluation particularly challenging.
For relatively straightforward tasks involving only a few to a few dozen operation sequences, datasets can scale to over one thousand instances \cite{10841395}. In contrast, for complex AMS IC tasks such as netlist or testbench generation, existing datasets contain only a few dozen highly intricate examples (Table~\ref{tab:ams-datasets}).
This highlights the need for benchmarks that closely reflect practical AMS design requirements, consistent with how practicing chip engineers learn from a limited number of carefully curated examples.

\begin{table}[h]
	\centering
	\small
	\begin{tabular}{c|c}
		\hline
		\textbf{Task}  & \textbf{Data Size} \\
		\hline
		Netlist Generation \cite{ho2024largelanguagemodelllm}  & 17 \\
		\hline
		Netlist Generation \cite{chen2025analogtesterlargelanguagemodelbased} & 24 \\
		\hline
		AMS-IO-Bench (Ours) & 30 \\
		\hline
	\end{tabular}
	\caption{Data sizes used in recent LLM-based agent studies for complex IC design tasks.}
	\label{tab:ams-datasets}
\end{table}

\subsection{Automation in I/O Ring Assembly}
Existing automation methods for I/O or padring construction require designers to perform extensive low-level specification through detailed configuration tables or metadata files \cite{YosysHQ_padring, chen2021padring, morita2015metadata, chen2010us7657858b2}, making these workflows nearly as labor-intensive as manual padring assembly.
These approaches are largely tailored for digital SoCs and offer limited support for AMS-specific needs such as power domain separation and custom analog pad cells; accommodating such constraints often requires direct modification of low-level scripts or tool code. 

None of these methods incorporates semantic interpretation or design-intent understanding, relying instead on explicitly specified parameters.
Their outputs are also limited to geometric files such as GDS that cannot be directly edited by designers.
Consequently, these methods operate at a different abstraction level and do not offer a directly comparable baseline for AMS I/O ring assembly.

\section{Method}

\subsection{Task Definition}

AMS I/O generation is defined as transforming human-provided pin planning specifications into production-ready schematics and layouts in EDA tools, as a constraint-driven engineering task focused on fulfilling design intent and ensuring compliance with all required design rules rather than performance or area optimization.

The input consists of pin planning specifications, typically expressed as tables or textual descriptions, defining I/O ring dimensions, pin names and ordering, and possibly additional requirements in natural language such as power domain separation or custom device usage.

The output is a complete set of production-ready schematics and layouts that can be directly integrated into the AMS IC design flow, as illustrated in Fig.~\ref{fig:Overview}. 
The generated results must conform to the intended geometry and pin arrangement, maintain strict electrical correspondence verified by Layout Versus Schematic (LVS) checks, and satisfy all physical design rules and manufacturability requirements verified by Design Rule Check (DRC).

\subsection{Agent Overview}
As illustrated in Fig.~\ref{fig:Overview}, AMS-IO-Agent is an LLM-based agent that bridges natural language design intent and executable EDA scripts through a structured pipeline of structured intent graph and intent graph adaptor. The agent consists of three key components:

\begin{itemize}
	\item \textbf{Design Intent Structuring}: Converts natural language or semi-structured specifications, such as pin lists with textual constraints, into a machine-readable graph that explicitly represents device configurations, spatial relationships, and electrical connections.
	
	\item \textbf{Intent graph adaptor}: Parses the intent graph to resolve constraints, perform geometric calculations, and derive implementation parameters, transforming structured intent into executable procedures.
	
	\item \textbf{Domain-specific knowledge base}: Provides design rules, device specifications, and layout conventions for constraint checking and design consistency with AMS I/O practices.
\end{itemize}

This layered architecture separates high-level intent reasoning from low-level implementation, allowing the LLM to focus on understanding design intent while deterministic modules ensure constraint resolution and reproducible script generation.

\subsection{Design Intent Structuring}
AMS I/O design often begins with informal specifications such as natural language descriptions, pin lists, or semi-structured spreadsheets, which contain essential design information but lack the structure required for automated processing. 
To address this gap, AMS-IO-Agent converts these inputs into a standardized \textit{intent graph}.

\begin{figure}[t]
	\centering
	\includegraphics[width=1\columnwidth]{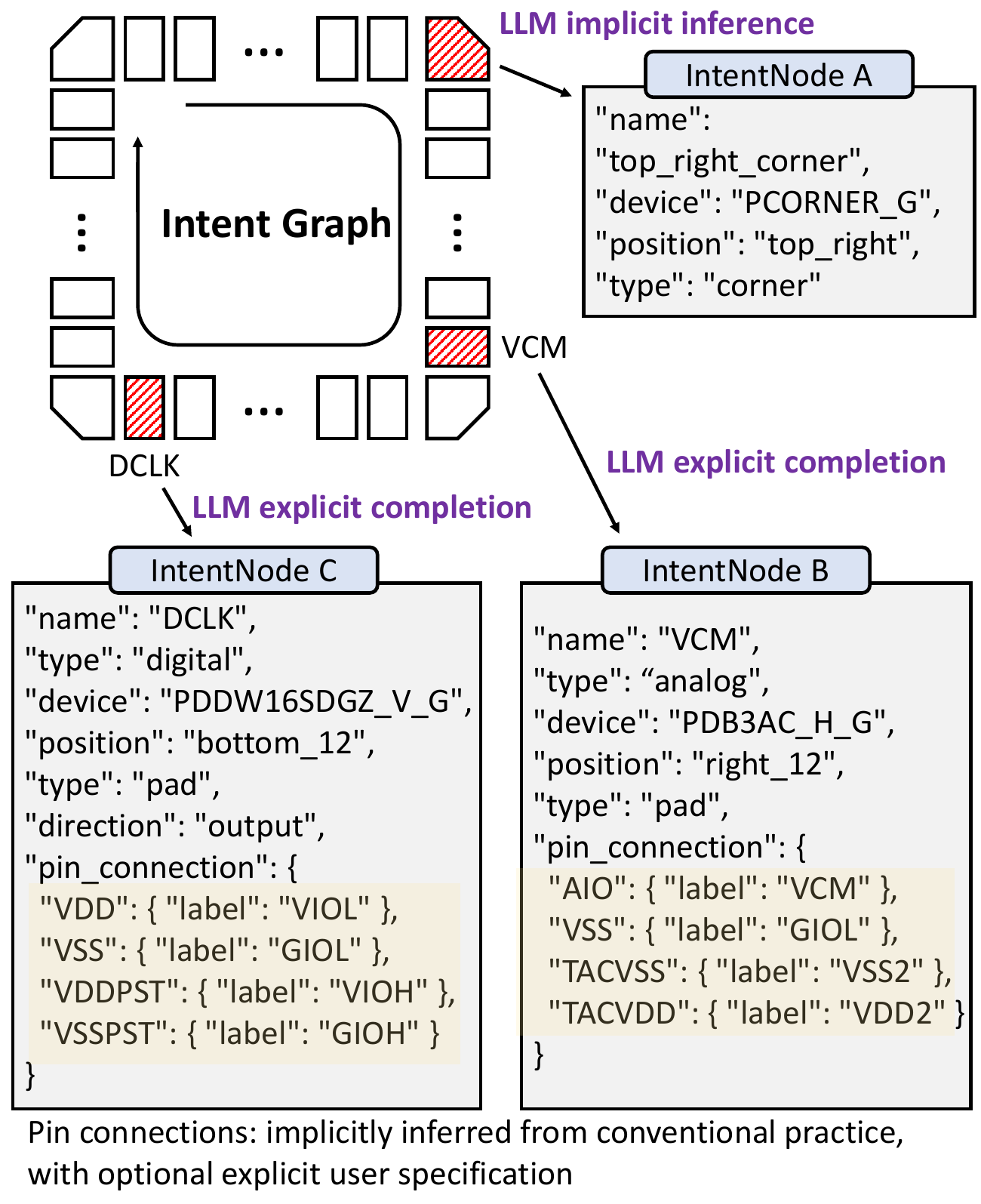} 
	\caption{Illustration of the intent graph and examples of its nodes.}
	\label{fig3}
\end{figure}

The intent graph is a JSON-based representation that models the I/O ring as a sequence of interconnected nodes, each representing a pad or corner cell with attributes such as name, device type, spatial position, direction, and pin connections (Figure~\ref{fig3}). 
Position encoding follows the I/O ring layout, preserving its physical organization.

The construction of the intent graph combines explicit completion and implicit inference.
Explicit completion is applied to pins whose names are provided in the specification. 
In practice, designers often use conventional abbreviations such as DCLK (digital clock), VCM (common-mode voltage) and VREFN (N-side of reference voltage). 
By leveraging these patterns through the domain-specific knowledge base, the agent can infer additional attributes, including signal type, device type, default direction, and pin connections.
These connections typically do not need to be specified explicitly by the user, although explicit user overrides are supported when required.
Implicit inference is used for elements that are not mentioned at all, such as corner cells, which are automatically inserted based on standard design conventions without any user input.

This representation is fundamentally different from a netlist and serves as the basis for middleware processing. 
While a netlist only describes circuit connectivity, the intent graph also captures spatial relationships, semantic context, and domain knowledge.
It can be directly understood by both humans and language models, and it can be efficiently parsed and processed by code. 
This dual accessibility makes it an effective interface between informal design inputs and automated implementation.

\subsection{Intent Graph Adaptor}

The Intent Graph Adaptor serves as a middleware layer that bridges the LLM-based agent and commercial EDA tools.
It supports the structured intent graph in completing EDA code generation and addresses the limitations of direct code generation for AMS I/O design.

Directly generating domain-specific language (DSL) scripts such as SKILL with an LLM is ineffective for producing correct and consistent outputs due to the lack of sufficient training data. 
Likewise, using these DSLs themselves to perform logical operations for parsing the intent graph is impractical because such languages are not designed for complex data manipulation.
To overcome these issues, the adaptor employs Python-based middleware for deterministic processing, including structured data parsing, constraint resolution, geometric calculations, and DSL script generation.

Although the middleware tools could, in principle, be generated by the LLM for every run, this approach suffers from low efficiency and poor robustness. 
Instead, implementing them as a reusable tool library allows the agent to invoke deterministic scripts for structured data parsing, constraint resolution, geometric calculations, and DSL script generation.
Upon receiving the intent graph, it extracts I/O instances with their attributes and computes precise cell coordinates based on I/O design rules.

For integration with commercial EDA tools, the adaptor generates SKILL scripts to create schematics and layouts in Cadence Virtuoso and csh scripts to invoke Siemens Calibre verification tools. 
Together, these components form a reproducible and verifiable bridge between the LLM-generated intent graph and tape-out-ready AMS I/O design flows.

\subsection{Domain-specific Knowledge Base}

The domain-specific knowledge base addresses a key challenge in AMS I/O design: design expertise is highly fragmented and often buried in informal natural-language documents such as training notes, reference guides, and internal design manuals. 
These materials encompass device selection practices, layout conventions, power domain rules, ESD protection requirements, naming conventions, common design techniques, and practical “know-how” accumulated from real projects.

To consolidate this expertise, AMS-IO-Agent employs a knowledge base built from the training materials and documented practices of a professional AMS IC design team of more than 10 experienced engineers, validated through over 50 successful tape-out cases. 
This knowledge base reflects the same materials used to train entry-level engineers, which can typically be mastered by engineers with an undergraduate background within one to three days, ensuring both its practicality and accessibility.

Instead of converting this knowledge into rigid rule code, AMS-IO-Agent organizes it into a lightweight repository of approximately 6k tokens. This repository serves both as a reference for human designers and as an in-context knowledge source for the LLM. Because of its compact size, the LLM can directly consume it without retrieval mechanisms or model fine-tuning.

By integrating verified expertise from actual design workflows into a single accessible knowledge base, the system enables both human and machine users to leverage the same authoritative engineering knowledge, combining human-readable documentation with automated, context-driven reasoning during intent interpretation and constraint resolution.

\section{Benchmark}

There is currently no standard evaluation framework or publicly available dataset for AMS I/O generation. 
To address this gap, we develop \textbf{AMS-IO-Bench}, the first benchmark suite dedicated to I/O ring generation for wirebond-packaged AMS IC chips.
Although other packaging methods such as flip-chip with solder bumps exist, wirebond remains the most commonly used approach for prototype verification chips. 
Because wirebond requires peripheral I/O rings along the chip boundary, it provides a clear and well-defined context for evaluating I/O planning, making it the most practical choice for our benchmark.

AMS-IO-Bench is derived from the I/O planning of 10 real tape-out projects collected over the past 5 years.
From these projects, we construct a benchmark of 30 cases by simplifying, augmenting, and transforming the original designs, ensuring that each case preserves the core constraints and design patterns of production flows. 
This approach enables both realism and reproducibility.
Each benchmark instance provides a structured pad location list specifying signal assignments, power domains, and routing hints.

The benchmark is organized into three difficulty levels, as shown in table \ref{tab:ams-io-bench-difficulty}.

\begin{table}[h]
	\centering
	\small
	\begin{tabular}{l c c p{3cm}}
		\hline
		\textbf{Level} & \textbf{Number} & \textbf{Realism} & \textbf{Features} \\
		\hline
		Simple & 10 & Simplified & Small size, single domain \\
		\hline
		Medium & 10 & Default & Standard size, multiple domains \\
		\hline
		Hard & 10 & Complex & Large size, staggered, customized cells \\
		\hline
	\end{tabular}
	\caption{Three difficulty levels in AMS-IO-Bench.}
	\label{tab:ams-io-bench-difficulty}
\end{table}

\begin{itemize}
	\item \textbf{Simple.} Simple cases use a single signal domain, thereby removing the need for implicit reasoning about isolation and local ESD power supply.
	They are derived from typical AMS IC designs through simplification. 
	
	\item \textbf{Medium.} Medium-difficulty cases reflect the default complexity of typical AMS IC designs. 
	Each case corresponds to a standard MPW chip outline of approximately 1mm × 1mm with a single-row I/O ring. 
	The I/O ring is partitioned into multiple power domains, including digital and analog, each governed by distinct device types, design rules, filler cells, and isolation cells. 
	Solving these cases requires the agent to reason over domain knowledge and contextual constraints, closely matching the real-world challenges encountered in tape-out projects.
	
	\item \textbf{Hard.} Hard cases represent advanced scenarios observed in complex tape-outs. 
	They include dual-row or partially dual-row I/O rings (staggered pads), chips with enlarged outlines (1.5× to 2× the default size or larger), custom I/O cells (e.g., analog I/O cells with reduced ESD capacitance), and designs with highly specialized power domain configurations such as localized ESD power delivery. These cases stress the adaptability of agents to highly customized design requirements.
\end{itemize}

By focusing on real-world constraints and vertical domain specificity, AMS-IO-Bench establishes a practical and scalable platform for evaluating AMS I/O design agents under realistic chip development conditions.

\begin{table*}[t]
	\centering
	\small
	\begin{tabular}{l|c|c|c|c|c|c|c}
		\hline
		\textbf{Method} & \textbf{IG (\%)} & \textbf{Shape (\%)} & \textbf{DRC (\%)} &
		\textbf{LVS (\%)} & \textbf{DRC+LVS (\%)} & \textbf{Time (min)} &
		\textbf{Token (k)} \\
		\hline
		Human & 100 & 100 & 100 & 100 & 100 & $\approx 480$ & -- \\
		\hline
		LLM (GPT-4o) & 0 & 0 & 0 & 0 & 0 & 0.2 & 1k \\
		\hline
		AMS-IO-Agent (GPT-4o) & 100 & 100 & 76.67 & 66.67 & 63.33 & 4.1 & 160k \\
		\hline
		AMS-IO-Agent (Claude-3.7) & 100 & 100 & 93.33 & 76.67 & 76.67 & 4.2 & 96k \\
		\hline
		AMS-IO-Agent (DeepSeek-V3) & 100 & 100 & 93.33 & 76.67 & 76.67 & 5.1 & 105k \\
		\hline
	\end{tabular}
	\caption{AMS-IO-Agent achieves perfect intent translation and shape validity, high DRC/LVS pass rates, and reduces design time from hours to minutes, outperforming a baseline LLM while remaining far more efficient than expert manual design.}
	\label{tab:main-results}
\end{table*}

\begin{table*}[t]
	\centering
	\footnotesize
	\setlength{\tabcolsep}{3.5pt}
	\renewcommand{\arraystretch}{1.2}
	
	\begin{tabular}{c c c | c c c c c | c c}
		\hline
		\textbf{KB} & \textbf{IG} & \textbf{Adaptor} &
		\textbf{IG (\%)} &
		\textbf{Shape (\%)} &
		\textbf{DRC (\%)} &
		\textbf{LVS (\%)} &
		\textbf{DRC+LVS (\%)} &
		\textbf{Time (min)} &
		\textbf{Token (k)} \\
		\hline
		
		$\times$ & $\times$ & $\times$ 
		& -- & 0 & 0 & 0 & 0 
		& 0.2 & 1 \\
		
		\checkmark & $\times$ & $\times^{\dagger}$ 
		& -- & 0 & 0 & 0 & 0 
		& 16.2 & 1098 \\
		
		\checkmark & $\times$ & \checkmark$^{\dagger\dagger}$
		& -- & 100 & 20.00 & 0 & 0 
		& 27.4 & 2036 \\
		
		\checkmark & \checkmark & $\times$ 
		& 100 & 0 & 0 & 0 & 0 
		& 2.3$^{\dagger\dagger\dagger}$ & 18 \\
		
		\checkmark & \checkmark & \checkmark
		& 100 & 100 & 93.33 & 76.67 & 76.67 
		& 5.1 & 105 \\
		\hline
		
		\multicolumn{10}{l}{\footnotesize $\dagger$\,LLM directly generates SKILL code without structured intent reasoning.}\\
		\multicolumn{10}{l}{\footnotesize $\dagger\dagger$\,LLM generates Python code for SKILL generation without structured intent reasoning.}\\
		\multicolumn{10}{l}{\footnotesize $\dagger\dagger\dagger$\,Reduced runtime because LVS/DRC tools cannot be invoked.}\\
	\end{tabular}
	
	\caption{Ablation results using DeepSeek-V3 show that both the knowledge base and structured intent adaptor are essential: removing either component eliminates signoff correctness.}
	\label{tab:component-ablation}
\end{table*}

\section{Experiments}
\subsection{Setup}

We implement and evaluate the proposed AMS-IO-Agent using the Python-based automation framework smolagents. 
The backbone language model is accessed via API. 
The agent runs on a workstation and communicates with chip-design server workstations via SSH and socket connections.

The tools invoked by the agent are implemented in Python. 
For integration with commercial EDA toolchains, the agent receives design intent descriptions as input, generates SKILL scripts for schematic and layout creation, and executes them in Cadence Virtuoso. 
Layout verification and physical rule checks are performed by invoking Siemens Calibre through csh scripts.
The evaluation follows the AMS-IO-Benchmark methodology described earlier.

\subsection{Evaluation Metrics}
We evaluate AMS-IO-Agent using a five-stage pipeline covering the entire process from intent interpretation to layout verification. Each metric targets a distinct stage, enabling systematic diagnosis and quantitative quality assessment.

\textbf{Metric 1: Intent Graph Pass Rate} checks whether the agent correctly converts natural language specifications into valid Intent Graphs, including proper pad naming, device type assignment, and attribute completion. Failures indicate misunderstanding of design requirements.

\textbf{Metric 2: Shape Score} quantifies layout similarity using a Vision-Language Model (VLM), which performs a binary evaluation (pass or fail) based on structural alignment with the reference. It captures visual and topological correctness beyond rule-based evaluation.

\textbf{Metric 3: DRC Pass Rate} reports the percentage of layouts passing all foundry-specified design rules (e.g., spacing, width, enclosure), reflecting physical manufacturability.

\textbf{Metric 4: LVS Pass Rate} checks electrical equivalence between schematic and layout, ensuring correct connectivity without opens, shorts, or mismatches.

\textbf{Metric 5: DRC+LVS Pass Rate} represents production-ready quality and serves as the overall effectiveness metric.

Together, these metrics provide a comprehensive evaluation framework spanning high-level intent interpretation and low-level physical validation.

\begin{figure*}[t]
	\centering
	\includegraphics[width=1\textwidth]{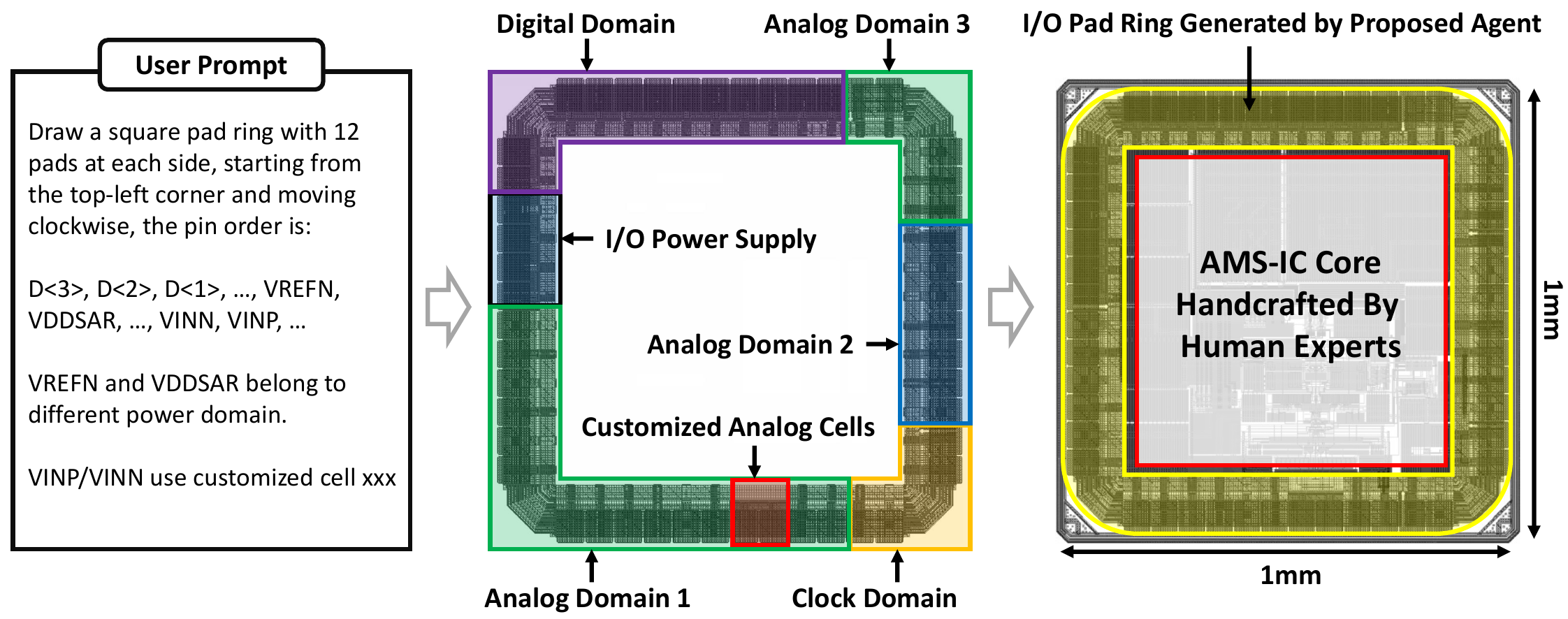} 
	\caption{Application of the proposed agent system to a real AMS tape-out I/O ring design.}
	\label{fig:case-study}
\end{figure*}

\subsection{Main Results}

We evaluate AMS-IO-Agent on AMS-IO-Bench, with results summarized in Table~\ref{tab:main-results}.
Across different backbone models, AMS-IO-Agent achieves 100\% intent graph translation and shape validity, while maintaining high DRC and LVS pass rates. Even for full DRC+LVS signoff, the agent reaches up to 76.7\% (23 out of 30 cases), demonstrating substantial progress toward production-ready layout generation with minimal human effort.
Compared to manual design, which typically requires around 480 minutes per task, the agent reduces turnaround time to only a few minutes and significantly outperforms baseline LLMs, with token usage remaining below 200k, making it practical for industrial deployment.
These results show that AMS-IO-Agent delivers a major efficiency boost while substantially improving functional design correctness over baseline methods.

\subsection{Error Handling}

Given the high cost (greater than \$10k per mm\textsuperscript{2}) and long fabrication time (about three months) of tape-out, human engineers primarily serve as reviewers. 
Infeasible cases detected during execution or through DRC and LVS can be resolved within minutes via prompt refinement or minor EDA edits, illustrating the advantage of generating editable intermediate deliverables rather than fixed geometric outputs.

\subsection{Ablation}

We further conduct an ablation study to assess the contribution of each component of AMS-IO-Agent, as shown in Table~\ref{tab:component-ablation}. 
Human time was estimated from interviews with 16 PhD level designers. 
Without the knowledge base (KB), intent graph (IG), or adaptor, the agent completely fails to generate valid outputs, highlighting the necessity of combining all three components. Using only the knowledge base enables some basic shape assembly but results in low DRC and no LVS signoff, indicating that design rules alone are insufficient without structured reasoning. Similarly, employing the intent graph without the adaptor allows the agent to generate syntactically valid intent representations but fails to translate them into signoff-quality implementations.
Only the full configuration, integrating KB, IG, and the adaptor, achieves 100\% in intent graph, shape, 93.33\% in DRC, and 76.67\% in DRC+LVS, while reducing runtime to 5.1 minutes with modest token usage (105k). These results confirm that the three components are complementary and jointly essential for practical AMS I/O design automation.

Table~\ref{tab:difficulty-validation} shows results across difficulty levels. All models succeed on simple cases, but their performance declines on medium and hard cases, validating that the benchmark’s difficulty scaling reflects real design complexity.

\begin{table}[ht]
	\centering
	\small
	\begin{tabular}{c|c|c|c}
		\hline
		\textbf{Model} & \textbf{Simple} & \textbf{Medium} & \textbf{Hard} \\
		\hline
		GPT-4o      & 10/10 & 7/10  & 2/10  \\
		\hline
		Claude-3.7  & 10/10 & 9/10  & 4/10  \\
		\hline
		Deepseek-V3 & 10/10 & 10/10 & 3/10  \\
		\hline
	\end{tabular}
	\caption{DRC+LVS pass rates of different models across difficulty levels in AMS-IO-Bench.}
	\label{tab:difficulty-validation}
\end{table}

\subsection{Case Study}
To demonstrate the real-world applicability of AMS-IO-Agent, we conducted a case study on a prototype tape-out project implemented in 28-nm CMOS technology, involving a 1 mm × 1 mm wirebond-packaged mixed-signal IC with 48 I/O pads (12 per side) and multiple power and signal domains. 
As illustrated in Figure~\ref{fig:case-study}, the inner AMS core was implemented by two human designers, while the surrounding I/O ring was generated by our agent.
This clear division of labor highlights the agent’s seamless integration into established design workflows. 

During the design process, a major pin-order change was introduced that required sliding and reordering many pads. 
While such a change would normally necessitate extensive manual redrawing, the agent regenerated a fully updated and verified layout within minutes.
The resulting I/O ring matched expert-designed quality and was readily adopted by the design team.

The final design, combining a manually crafted AMS core with the agent-generated I/O ring, passed LVS and DRC and was successfully fabricated.
Silicon measurements confirmed correct functionality.
This case study demonstrates that AMS-IO-Agent can serve as a production-ready module within real tape-out workflows, enabling true human-agent collaborative design while significantly reducing manual rework effort during late-stage iterations.

\section{Conclusion}

In this work, we introduced \textit{AMS-IO-Agent}, a domain-specialized LLM-based agent for structure-aware AMS I/O generation, together with \textit{AMS-IO-Bench} for systematic evaluation.
By combining a curated domain knowledge base with structured intent representation, our approach translates informal pin-planning specifications into production-ready EDA workflows.

Experiments on AMS-IO-Bench show that AMS-IO-Agent achieves 100\% intent graph correctness, 100\% layout shape validity, and over 70\% DRC+LVS signoff pass rate, reducing design turnaround time from days to mere minutes. 
Moreover, a prototype tape-out confirms the agent’s capability to seamlessly collaborate with human engineers, adapt to late-stage pin-order changes, and deliver signoff-quality I/O rings suitable for real-world chip design flows.

These results establish the feasibility of deploying LLM-based agents for AMS IC I/O automation and provide a concrete foundation for human-agent co-design. 
The methodology applies broadly to AMS layout tasks that exhibit regular structures and semantically meaningful signal naming, and can be adapted to other technology nodes or packaging styles by substituting the knowledge base and execution adaptor.
Future work will extend this approach to more complex AMS design tasks and enable deeper integration with downstream verification and routing tools.

\noindent
\textbf{Limitation and Social Impact}
Our approach focuses on wirebond-packaged AMS I/O rings and relies on a domain knowledge base tailored to specific design conventions, which may limit generalization to other packaging types or foundry rules. While AMS-IO-Agent reduces manual workload, it cannot fully replace expert review, especially for highly customized or unconventional designs.

From a social perspective, our method can boost productivity and lower barriers for AMS IC design, but may also shift the demand for certain engineering skills. Responsible use and human oversight remain essential to ensure safety and reliability in real-world chip development.

\section{Acknowledgments}
This work was supported by National Science Foundation of China (NSFC) Young Scholar Basic Research Program for doctoral students (Grant 624B2081).

\bibliography{aaai2026}

\end{document}